\documentclass{ieeeaccess}
\usepackage[T1]{fontenc}
\usepackage[utf8]{inputenc}
\usepackage{cite}
\usepackage{amsmath,amssymb,amsfonts}
\usepackage{algorithmic}
\usepackage{textcomp}
\usepackage{graphicx}
\usepackage{comment}
\usepackage{booktabs} 
\usepackage{multirow} 
\usepackage{array} 
\def\BibTeX{{\rm B\kern-.05em{\sc i\kern-.025em b}\kern-.08em
    T\kern-.1667em\lower.7ex\hbox{E}\kern-.125emX}}
\usepackage{aeguill}
\usepackage[colorlinks=true, linkcolor=blue, urlcolor=blue, citecolor=blue]{hyperref}
\makeatletter
\newcommand\blfootnote[1]{%
  \begingroup
    \renewcommand\thefootnote{}\footnote{#1}%
    \addtocounter{footnote}{-1}%
  \endgroup
}

\begin{document}
\history{Date of publication xxxx 00, 0000, date of current version xxxx 00, 0000.This work has been submitted to the IEEE for possible publication. Copyright may be transferred without notice, after which this version may no longer be accessible.\\
}
\doi{10.1109/ACCESS.2017.DOI}

\title{Large Language Model-Driven Dynamic Assessment of Grammatical Accuracy in English Language Learner Writing}

\author{\uppercase{Timur Jaganov}\authorrefmark{1},
\uppercase{John Blake\authorrefmark{1}\IEEEmembership{Member, IEEE}, \uppercase{Juli\'an Villegas}\authorrefmark{1}\IEEEmembership{Senior Member, IEEE}, and Nicholas Carr}\authorrefmark{1}
}
\address[1]{University of Aizu, Aizuwakamatsu, Japan (e-mail: \{m5281502, jblake, julian, carrnick\}@u-aizu.ac.jp)}

\tfootnote{This work was supported by the Japan Society for the Promotion of Science (JSPS) Grant-in-Aid for Scientific Research (Kakenhi), Grant Number 23K00656.}

\markboth
{Jaganov \headeretal: Large Language Model-Driven Dynamic Assessment}
{Jaganov \headeretal: Large Language Model-Driven Dynamic Assessment}

\corresp{Corresponding author: Nicholas Carr (e-mail: carrnick@u-aizu.ac.jp).}

\begin{abstract}
This study investigates the potential for Large Language Models (LLMs) to scale-up Dynamic Assessment (DA). To facilitate such an investigation, we first developed DynaWrite—a modular, microservices-based grammatical tutoring application which supports multiple LLMs to generate dynamic feedback to learners of English. Initial testing of 21 LLMs, revealed GPT-4o and neural chat to have the most potential to scale-up DA in the language learning classroom. Further testing of these two candidates found both models performed similarly in their ability to accurately identify grammatical errors in user sentences. However, GPT-4o consistently outperformed neural chat in the quality of its DA by generating clear, consistent, and progressively explicit hints. Real-time responsiveness and system stability were also confirmed through detailed performance testing, with GPT-4o exhibiting sufficient speed and stability. This study shows that LLMs can be used to scale-up dynamic assessment and thus enable dynamic assessment to be delivered to larger groups than possible in traditional teacher-learner settings.
\end{abstract}


\titlepgskip=-15pt

\maketitle
\blfootnote{%
  This work has been submitted to the IEEE for possible publication.%
   Copyright may be transferred without notice, after which this version %
    may no longer be accessible.%
}
\section{Introduction}
\label{sec:introduction}
Conventional approaches to assessment tend to require learners complete tasks independently, without any external assistance from the teacher. 
In language learning classrooms, such assessments typically include students independently completing receptive or productive tasks. 
Without diminishing the role of such assessments, they only inform teachers of the language features a student has mastered and offer little insight into the features a learner is in the process of mastering. 
\subsection{Dynamic assessment}
\label{subsec:dynamic_assessment}
Dynamic assessment (DA) differs from conventional assessment due to its inclusion of graduated external assistance when a learner encounters difficulties \cite{poehner2018probing}. 
Graduated assistance refers to the learner receiving dynamic feedback which begins with implicit hints and increases in explicitness as per learner needs \cite{aljaafreh1994negative}. 
Dynamically assessing writing in the language classroom usually entails learners receiving dynamic feedback to help them identify and correct grammatical errors in their text. 
One advantage of this approach is that it reveals which language features are in the process of maturing \cite{poehner2018probing}. 
In other words, when only implicit assistance is required, the language feature is in the process of being mastered. 
Another potential advantage is that dynamic feedback may develop knowledge which helps learners use language with greater control in subsequent situations \cite{poehner2018probing}. 

One obstacle for teachers wishing to harness the benefits of DA is its lack of scalability—it is difficult to deliver dynamic feedback to large groups as it must be tailored to each individual learner's needs. 
While there have been advancements in the ability of automated written error detectors to provide feedback to learners due to the shift from rule-based transformations to sophisticated machine learning methods, they still lack the capability to provide dynamic feedback. 
It is this niche that this research explores by investigating if LLMs can be utilized to design tool which provides dynamic feedback. 

\subsection{Feedback in Language Learner's Writing}
\label{subsec:importance-wcf}
The provision of feedback in response to an error that has occurred in the writing of a language learner is referred to as written corrective feedback (WCF) \cite{bitchener2016written}. 
Despite extensive research on the potential benefits of WCF for language learners \cite{liu2015methodological}, there continues to be debate concerning its pedagogical value, the efficacy of different types of WCF and modes of delivery. 
Nonetheless, WCF continues to be extensively used in language learning classrooms, with both institutions \cite{alshahrani2014investigating} and students expecting teachers to provide feedback \cite{lee2004error}. 
One line of research emerging from the extant literature that shows promise is the oral delivery of dynamic feedback, i.e., the teacher orally provides hints on errors in face-to-face environments to individuals or small groups, with the hints increasing in explicitness as per learner's real-time needs (see Table~\ref{tab:feedback_types} for an example). 
\begin{table}[htb]
    \caption{Example of Dynamic Feedback (adapted from \cite{Poehner2009-kj})}
    \label{tab:feedback_types}
    \centering
    \begin{tabular}{@{}p{2cm}p{6cm}@{}}
        \toprule
        \textbf{Level} & \textbf{Feedback Type} \\
        \midrule
        \multirow{3}{3cm}{Most Implicit} 
            & 1. Ask student if there is anything wrong with this sentence? \\
            & 2. Repeat phrase in the sentence which contains the error \\
            & 3. Point out the incorrect word(s) \\[3pt]
        Most Explicit & 4. Provide correct form and explain why error occurred \\
        \bottomrule
    \end{tabular}
    
\end{table}

While this approach has often been found to be effective \cite{nassaji2000vygotskian,rassaei2021effects}, its delivery is impractical in large classes due to its time-intensive nature \cite{erlam2013oral}, thus further highlighting the need for scalable automated process.
\subsection{Theoretical Underpinnings}
\label{subsec:sociocultural-theory}
DA is underpinned by tenets of sociocultural theory (SCT). A detailed description of SCT is beyond the scope of this paper. Accordingly, we refer interested readers to seminal works, such as  \cite{vygotsky1987collected, lantolf2014sociocultural}. However, there are two key tenets relevant for our research.  First, symbolic tools, such as language, mediate our mental functions \cite{Vygot}. In other words, language acts as a tool which helps us gain mastery over our thinking \cite{lantolf2014sociocultural}. When completing a task independently without assistance, we mediate ourselves by drawing already matured functions. However, when completing a task with the assistance of external resources, such as assistance from a teacher, our functioning is mediated by those resources \cite{poehner2018probing}. The second is that it is more meaningful to understand what a learner can achieve when additional mediation, such as teacher hints, is available rather than what an individual can achieve when working independently  \cite{vygotsky19989collected5}. Understanding this uncovers what functions are in the process of maturing, and thus provides crucial insights to guide and shape subsequent instruction  \cite{vygotsky2012thought}.  

As previously noted, to realistically apply DA in whole-class instruction, scalable solutions are required. Computerized Dynamic Assessment (CDA) offers a promising way forward by delivering personalized mediation at scale. To date, most CDA systems have primarily focused on delivering DA in receptive skill domains such as reading and listening \cite{randall2023development}. However, there remains a need in both research and practice for CDA systems that can emulate the individualized mediation traditionally provided by human teachers during more productive language tasks such as speaking and writing. 

Our research investigates whether LLMs can deliver DA of grammatical accuracy in extended learner writing by:
(1) accurately identifying grammatical errors,
(2) providing graduated, pedagogically appropriate feedback, and
(3) operating effectively within a scalable digital platform.

\subsection{Overview}
In what follows, we first provide an overview of related works of CDA in Section II. In Section III we present an in-depth description of the design of our proposed system, DynaWrite. This is followed by an overview of the iterative process utilized in the development of DynaWrite in Section IV. Section V details the evaluation of the tool in terms of usability and accuracy. A discussion of these results and a conclusion are then provided in Sections VI and VII respectively.   

\section{Related Works}
\label{sec:related-works}
\subsection{Automated Corrective Feedback}
Recent research 
\cite{burstein2004automated,leacock2009user,ding2024automated}
has shown increasing interest in computerized grammatical error identification and correction systems for both improving the grammatical accuracy of first language (L1) users of English and second language (L2) users of English. A key limitation of these systems is that they are primarily designed with the goal of improving the quality of the text and not on developing the user's knowledge to prevent the same errors recurring in writing performed at a later time.  

Automated Writing Evaluation (AWE) tools have been developed to provide instant feedback on grammatical errors. 
Although initially developed as a proofing tool, Grammarly is widely used in L2 classrooms, and offers real-time feedback on grammar, spelling, and punctuation errors \cite{ding2024automated}. Criterion\textsuperscript{\textregistered}, developed by the Educational Testing Service, analyzes students' writing grammatically and provides sentence-level error feedback \cite{chapelle2015validity}.
CyWrite \cite{chukharev2019empowering}, a grammatical analyzer developed for undergraduate students with English as their L2, has shown better performance in detecting certain grammatical errors compared to Criterion \cite{feng2016automated}.
CyWrite adopts a novel approach to automated writing evaluation, which focuses on both the drafting process and the final product. This is achieved through real-time processing of keystroke logs, which record the false starts, corrections and revisions that learners tend to make while writing.

While acknowledging the advancements of these tools, they lack the ability to offer graduated assistance. Consequently, the potential benefits of DA cannot be harnessed with these tools. 

\subsection{Computerized Dynamic Assessment Tools}

To meet the demands of scalability, various CDA systems have been successfully designed, with their implementation showing many of the benefits of DA are retained when scaling up to CDA, for example  \cite{ai2017providing,zhang2019measuring,leontjev2016dynamic,mehri2021diagnosing,poehner2013bringing,poehner2015computerized,randall2023development,yang2017assessing}. This body of work has utilized CDA for receptive skills such as listening, for example \cite{mehri2021diagnosing,zhang2019measuring}, reading  \cite{yang2017assessing}, both listening and reading  \cite{poehner2015computerized} and word derivational knowledge \cite{leontjev2016dynamic}. These studies found some benefits of DA carry over to CDA environments, with the main benefit being that CDA tools can provide detailed and insightful diagnostic information on the linguistic functions which are in the process of maturing and help inform subsequent learning materials. While some of these systems have included the construct of grammar, it has been limited to the context of understanding grammar in a listening or reading task and not included the productive use of language. One notable system which begins to fill this gap is described in Randall and Urbanski\cite{randall2023development}. While acknowledging the advancement their system makes, it is built around users selecting the missing word from a sentence in multiple choice questions to complete a sentence or constructing sentences by putting words in order. In other words, the system does not provide CDA in a context which allows the learner to write their own original texts.

Blake \cite{blake2020genre} developed an online genre-specific error detection tool that provides multimodal feedback on learner's freely written texts. The feedback focuses on the accuracy of a limited set of grammatical features, and four other types of errors, namely brevity, clarity, objectivity, and formality. Feedback is provided at two levels of explicitness, beginning with very implicit feedback and learners then having the option to access feedback at the higher level. While the system does not provide the graduation normally associated with DA, it is a step towards a system which approximates CDA due to two levels of feedback being available. Moreover, the tool does not address accuracy errors detectable by more sophisticated AI-powered correction tools, such as Grammarly. 

Nicholas et al. \cite{nicholas2024effectiveness,nicholas2024profiling,blake2025computerized}
developed a CDA tool that focuses on pragmatic errors in email writing. To the best of our knowledge, their system 
was the first dynamic language assessment system to parse complete texts. Students submit drafts of emails and receive diagnostic feedback whenever instances of pragmatic failure are detected. The tool, however, is not designed to provide feedback on non-pragmatic errors, such as grammatical inaccuracies.

It is important to note that neither of these systems harnessed LLMs. To investigate the potential of LLMs being utilized in CDA, Blake \cite{blake2024unleashing} described a small-scale pilot study using an LLM as a dynamic assessment tutor, noting that, at that time, ChatGPT \cite{openai2022chatgpt} was unable to provide graded feedback with increasing levels of explicitness in a manner comparable to that of a human. 

Early approaches to CDA relied on rule-based systems that followed predefined templates to identify learner errors. While these methods provided consistent feedback, they lacked the ability to deal with learner-specific responses, especially in complex, open-ended tasks, such as extended writing \cite{choi2024exploring}. LLMs, however, have the potential to transform CDA. LLMs facilitate real-time, context-sensitive feedback generation through probabilistic reasoning rather than relying on fixed rules or narrowly supervised datasets, enabling individualized tailored feedback, geared to the learners’ evolving understanding and error patterns \cite{morris2024automated}. Unlike earlier systems, LLM-based assessment tools can interpret a wider range of learner input and adjust their responses accordingly, and as such potentially more closely approximating in-person teacher-student interactions.  
Therefore, this research seeks to more fully investigate the potential of LLMs being used as a means to implement CDA that focuses on the grammatical accuracy within extended written texts.

\section{System Design}
\label{sec:system_design}

\subsection{Overview and Objectives}
The web application DynaWrite was designed to address two main tasks: (1) administering dynamic assessment on grammatical errors occurring in texts written by Japanese learners of English, and (2) evaluating the capabilities of LLMs in real-world educational settings.

The primary pedagogical aim of task one is to enhance learning outcomes by providing tailored, iterative feedback. DynaWrite delivers dynamic assessment by offering increasingly explicit guidance with each attempt, encouraging learners to resolve errors independently. 
Table~\ref{tab:feedback-levels} is an adaption of \cite{poehner2018probing} which shows examples of the type of feedback that is provided at each level of explicitness. 
If the learner is unable to correct the error after the maximum number of attempts, the system presents the correct form, balancing support with learner autonomy.

Additional design objectives for task one include ensuring modularity and scalability. Modularity allows for the flexible integration of new error types and feedback strategies without restructuring the system, supporting future expansion and customization. 
Scalability ensures the system can accommodate growing user numbers and extended usage without performance degradation, making it suitable for deployment in institutional or nationwide educational contexts.

The second task involves the evaluation of LLMs embedded in the application. Here, the goal is to identify models that offer an optimal trade-off between performance and accuracy. 
Specifically, DynaWrite compares multiple LLM backends to determine which model delivers the most accurate grammatical error identification and correction while maintaining real-time responsiveness and low computational cost. 
This ensures that the selected model is both pedagogically effective and operationally viable for large-scale use.



\begin{table}[htb]
\caption{Feedback Levels, Types and Their Descriptions.}
\label{tab:feedback-levels}
\centering
\begin{tabular}{@{}p{16pt}p{66pt}p{118pt}@{}}
\toprule
\textbf{Level}  & \textbf{Type} & \textbf{Description} \\
\midrule
1 & Implicit Hint & Indicates the presence of an error without specifying details. \\
2 & Probing Question & Asks a probing question to help the user identify the location of the error. \\
3 & Error location & Identifies the problematic section of the sentence  \\
4 & Explicit Correction & Suggests a correct version of the sentence and provides an explanation of the error. \\
\bottomrule
\end{tabular}
\end{table}

DynaWrite offers researchers robust tools to analyze the performance, efficiency, and precision of LLMs in real-world writing tasks.
Its scalable architecture allows for the seamless addition or replacement of LLMs, minimizing the need for extensive reconfiguration and ensuring adaptability for future developments.

The system harnesses a microservices-based architecture, powered by \texttt{Apache\_Kafka}, which supports dynamic model integration, enabling  LLMs to be added, removed, or replaced without disrupting the workflow.
This flexibility ensures that DynaWrite remains a cutting-edge tool for education and research.

\subsection{System Architecture}
\begin{table}
\caption{Microservices in \texttt{DynaWrite}}
\label{tab:microservices}
\setlength{\tabcolsep}{3pt}
\begin{tabular}{@{}p{85pt}p{140pt}@{}}
\toprule
\textbf{Microservice} & \textbf{Description} \\ 
\midrule
\texttt{REST\_API} & Provides core backend functionality for processing user requests, authenticating users, and managing interaction with \texttt{Apache\_Kafka}.\\ 
\texttt{DynaWrite\_Consumer} & Listens to \texttt{Apache\_Kafka} messages, processes \texttt{NLP} responses, stores results in \texttt{PostgreSQL}, and sends updates to \texttt{Centrifugo}.\\ 
\texttt{Ollama\_Consumer} & Handles requests for \texttt{LLM} in the \texttt{Ollama} framework, such as \texttt{Llama3} and \texttt{Gemini}.\\ 
\texttt{OpenAI\_Consumer} & Manages communication with the \texttt{OpenAI ChatGPT API} for text analysis.\\ 
\texttt{Export\_Consumer} & Generates reports in \texttt{Excel} and \texttt{CSV} formats, detailing \texttt{LLM} performance and user interaction data.\\ 
\texttt{CPush\_Consumer} & Sends real-time updates to the frontend via \texttt{WebSocket} channels.\\ 
\bottomrule
\end{tabular}
\end{table}
DynaWrite 
uses a modular microservices architecture, making it scalable, reliable, and easy to integrate. Figure~\ref{fig:architecture} shows how users interact with different parts of the system.
Users work through a \texttt{VueJS}-powered front-end \cite{vuejs}, serviced using \texttt{Nginx} \cite{ma2022evaluation}, which connects to a \texttt{Go/Gin}-based \texttt{REST API} backend.
The \texttt{Apache Kafka} message broker \cite{apachekafka} facilitates communication between microservices responsible for \texttt{LLM}, reporting, and real-time updates.
The architecture uses \texttt{PostgreSQL} for data storage and \texttt{Keycloak} for authentication \cite{keycloak}.
Separate microservices connect \texttt{LLMs} like \texttt{GPT-4o-mini} \cite{openai_gpt4o_mini_2024} and \texttt{Llama3.2} \cite{meta_llama3_2_2024}. The system consists of six microservices, each with a distinct task.
Additional infrastructure components ensure stable and efficient operation of the system.

\Figure[t!](topskip=0pt, botskip=0pt, midskip=0pt)[width=\linewidth]{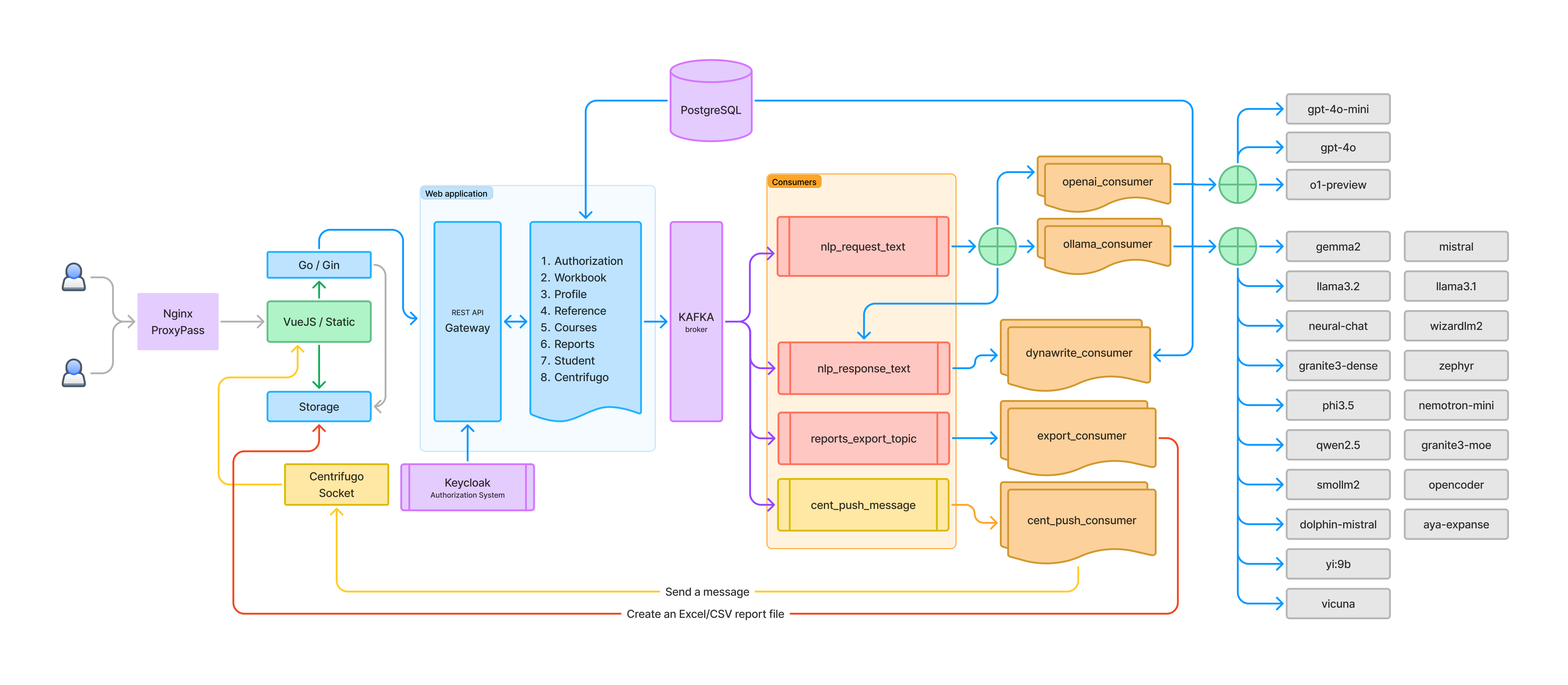}
{The architecture of the DynaWrite system.\label{fig:architecture}}


The six microservices in DynaWrite are summarized in Table~\ref{tab:microservices}, along with their respective roles. 
The architecture also includes the following infrastructure components:
\begin{itemize}
    \item \texttt{Nginx}: Served as a reverse proxy for the \texttt{VueJS} frontend and \texttt{Go} backend.
    \item \texttt{Apache Kafka}: Facilitates asynchronous communication between microservices through well-defined topics (See Table~\ref{tab:kafka_topics}).
    \item \texttt{PostgreSQL}: Stores user data, session logs, and \texttt{NLP} analytics in a relational schema.
    \item \texttt{VueJS}: Implements the user interface, which is compiled into static files served by \texttt{Nginx}.
    \item \texttt{Centrifugo}: Ensures responsive, real-time feedback for users interacting with the system.
    \item \texttt{Keycloak}: Manages user authentication and role-based access control.
\end{itemize}

\begin{table}
\caption{Kafka Topics Used in DynaWrite.}
\label{tab:kafka_topics}
\setlength{\tabcolsep}{3pt}
\begin{tabular}{@{}p{95pt}p{130pt}@{}}
\toprule
\textbf{Topic Name} & \textbf{Purpose} \\ 
\midrule
\texttt{nlp\_request\_text} & Sends text analysis requests to \texttt{NLP} microservices.\\ 
\texttt{nlp\_response\_text} & Receives error analysis and dynamic feedback from \texttt{LLM}.\\ 
\texttt{cent\_push\_message} & Delivers real-time feedback to the frontend.\\ 
\texttt{reports\_export\_topic} & Handles requests for generating analytical reports.\\ 
\bottomrule
\end{tabular}
\end{table}

\subsection{User-focused Functionalities}

The DynaWrite system provides a comprehensive set of functionalities. Some functionalities are targeted to specific user roles, such as report generation for various LLM models, course creation (which comprises different assignments for students packaged in a specific curriculum order), course completion tracking, and workbook functionality (allowing users to create personal working documents where they can edit their text with AI assistance independent of any lesson). 
Other functionalities that are shared across all user groups include authentication and role management, real-time updates, and accessibility. 
User authentication and role management is managed through \texttt{Keycloak}, ensuring secure access based on user roles, i.e. student, teacher, or researcher.
Real-time updates are powered by \texttt{Centrifugo}, ensuring instant feedback and seamless interaction between users and the system. Accessibility is supported through a responsive design that ensures compatibility across a range of devices, including desktops, tablets, and smartphones.

Targeted functionalities are designed to cater to three primary user groups: students, teachers, and researchers.
Each group benefits from tailored tools and features aimed at enhancing the user experience, improving learning outcomes, and supporting research.

\subsubsection{Student focus}

Students are the primary end-users of the system, utilizing DynaWrite as a learning platform to improve their English grammar in context.
Key functionalities include workbooks for text correction, progress tracking, and real-time feedback.

\paragraph{Workbook for Text Correction}
\begin{itemize}
    \item Students can type text  in the \textbf{Workbook} section (See Figure~\ref{fig:workbook_screenshot}).
    \item The system processes the text sentence by sentence, parsing for grammatical errors using selected NLP models.
    \item Feedback is provided through a \textbf{hierarchical hint system}, offering four levels of hints as shown in Table \ref{tab:feedback-levels}.
    \item Students can revise each sentence up to three times, progressively receiving more explicit feedback if errors persist. After the third attempt, a suggested sentence is provided, accompanied by a brief explanation of the error.
    \item Sentences corrected successfully are marked as \textbf{``completed''}, while the remainder are logged for further review.
\end{itemize}

\Figure[th](topskip=0pt, botskip=0pt, midskip=0pt)[width=.95\linewidth]{methods/screenshot-mobile.png}
{The DynaWrite Workbook mobile interface. This screenshot shows DynaWrite providing a Level 2 hint to help the learner revise a sentence flagged as grammatically incorrect. Key parameters like the LLM, temperature, and token settings are displayed for both transparency and research purposes.\label{fig:workbook_screenshot}}

\paragraph{Progress Tracking}
\begin{itemize}
    \item Students can view a history of their previous sessions, including the number of grammatical errors identified and corrected, the level of hint required to correct each sentence, and a summary of incorrect sentences.
    \item Progress indicators help students track their improvement over time (See Figure~\ref{fig:tracker_screenshot}).
\end{itemize}

\paragraph{Real-Time Feedback}
\begin{itemize}
    \item Real-time error analysis is powered by LLMs and delivered to students through the Centrifugo-based WebSocket system.
    \item Instant feedback enhances the learning experience by maintaining engagement and minimizing delays.
\end{itemize}

\Figure[h!](topskip=0pt, botskip=0pt, midskip=0pt)[width=.65\linewidth]
{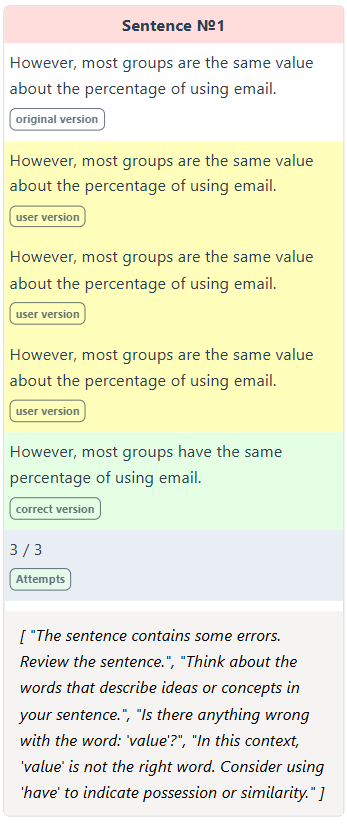}
{The DynaWrite Progress Tracker interface. This screenshot shows the original submitted sentence at top of the column. The versions with a yellow background are inaccurate user-submitted revisions while the version with a green background is either the users correct revision or the suggested correction if the user could not revise correctly in three attempts. The hints provided are given at the bottom of the column.\label{fig:tracker_screenshot}}

\subsubsection{Teacher focus}

Although the DynaWrite system adopts the role of tutor, teachers play a key role in guiding students and monitoring their progress. DynaWrite provides educators with the following functionalities: course and task management, monitoring student performance, and report generation.

\paragraph{Course and Task Management}
\begin{itemize}
    \item Teachers can create and assign courses consisting of various writing tasks. These tasks can range from simple essay question prompts to image descriptions, such as describing a graph or table.   
    \item Tasks are customizable with parameters such as selecting the type of writing task, and guidelines to display to students.
\end{itemize}

\paragraph{Monitoring Student Performance}
\begin{itemize}
    \item Teachers can access detailed reports on student performance, detailing the success rates for text corrections, the average number of attempts required per sentence, and the number of hints utilized at each level of explicitness.
\end{itemize}

\paragraph{Report Generation}
\begin{itemize}
\item DynaWrite can generate downloadable reports in \texttt{CSV} and \texttt{XLSX} formats for further analysis. 
\item Reports include details on, \textit{inter alia}, the LLM used for analysis, response times and accuracy metrics, and sentence-level summaries of student performance. These reports are particularly useful for identifying the language features which are in the process of maturing for learners and thus inform subsequent assessments and instruction. 
\end{itemize}

\subsubsection{Researcher focus}

DynaWrite is designed with modularity and advanced analytics, making it a valuable resource for researchers investigating the performance of LLMs in educational settings.
Its architecture allows for detailed evaluation, benchmarking, and data collection to support cutting-edge research.

One of the system’s key features is its capability for comparative analysis of LLMs.
Researchers can evaluate multiple models based on descriptive statistics such as mean response time of the model, standard deviation, percentiles, accuracy in detecting grammatical errors, and the rates of false positives and negatives.
The system provides visual insights through detailed graphs and charts (e.g., Figures~\ref{fig:accuracy}–\ref{fig:avg_response_time}), which enable a comprehensive understanding of model performance.

In addition to comparative analysis, DynaWrite supports performance benchmarking by allowing researchers to integrate new LLMs via Apache Kafka.
The system automatically collects and stores performance metrics, creating a structured dataset for in-depth evaluation and optimization of these models.

Another significant feature is its capability for dataset collection and annotation.
The system logs anonymized user interactions, including data on error correction attempts and the effectiveness of hierarchical hints.
These datasets are instrumental in training and fine-tuning future LLMs and are valuable for conducting broader linguistic research and understanding user behavior.

By combining these functionalities, DynaWrite offers researchers a powerful platform to explore the potential of LLMs and contribute to advancements in educational technology and computational linguistics. 

\section{System Evaluation}
This section addresses the three core components of the research question: (1) the accuracy of grammatical error identification, (2) the appropriacy of feedback generated for pedagogical use, and (3) the suitability of the platform for real-time, scalable deployment. Each of these aspects was systematically evaluated through targeted testing and user analysis. The following subsections present the key findings.

\subsection{Accuracy of Error Identification}
To evaluate the ability of DynaWrite to accurately detect grammatical errors in learner writing, four standard classification metrics were employed: true positives (TP), false positives (FP), true negatives (TN) and false negatives (FN). These metrics reflect cases where errors were correctly identified, instances where correct text was mistakenly flagged, where no errors occurred, and errors that were missed. 

The initial evaluation involved a set of 21 models, selected based on the following criteria: (1) model size under 6GB to ensure feasibility for local deployment, (2) established reputation or popularity within the NLP community, and (3) release within the previous 12 months to reflect current capabilities. The models used during testing are summarized in Table~\ref{tab:ollama_models_sorted}.

\begin{table}
\caption{Summary of Ollama Models Used in Testing (Sorted by Size)}
\label{tab:ollama_models_sorted}
\setlength{\tabcolsep}{3pt}
\begin{tabular}{|p{75pt}|p{25pt}|p{115pt}|}
\hline
\textbf{Model Name} & \textbf{Size (GB)} & \textbf{Description} \\ \hline
\texttt{gemma2} & 5.4 & An advanced model developed for handling complex linguistic tasks with high proficiency.\\ \hline
\texttt{aya-expanse} & 5.1 & A general-purpose LLM suitable for a wide range of language understanding and generation applications.\\ \hline
\texttt{yi:9b} & 5.0 & A large-scale language model designed for comprehensive understanding and generation capabilities.\\ \hline
\texttt{qwen2.5} & 4.7 & A language model series by Alibaba, pretrained on large-scale datasets, supporting extended context lengths.\\ \hline
\texttt{opencoder} & 4.7 & An encoder model optimized for processing and understanding input data in various NLP tasks.\\ \hline
\texttt{llama3.1} & 4.7 & A versatile model for natural language processing tasks, including text generation and summarization.\\ \hline
\texttt{zephyr} & 4.1 & A fine-tuned Mistral model for assistant-like tasks, available in various parameter sizes.\\ \hline
\texttt{neural-chat} & 4.1 & A conversational AI model designed for generating human-like dialogue responses.\\ \hline
\texttt{mistral} & 4.1 & A language model known for its balance between performance and efficiency in various NLP tasks.\\ \hline
\texttt{dolphin-mistral} & 4.1 & A variant of the Mistral model, fine-tuned for specialized language processing tasks.\\ \hline
\texttt{wizardlm2} & 4.1 & A language model tailored for advanced natural language understanding and generation.\\ \hline
\texttt{vicuna} & 3.8 & A general-use chat model based on Llama and Llama 2, optimized for conversational applications.\\ \hline
\texttt{nemotron-mini} & 2.7 & A compact model engineered for specific NLP applications requiring lower computational resources.\\ \hline
\texttt{phi3.5} & 2.2 & A language model designed for efficient natural language understanding and generation.\\ \hline
\texttt{granite3-moe} & 2.1 & A mixture of experts model aimed at enhancing grammatical accuracy in text processing.\\ \hline
\texttt{llama3.2} & 2.0 & A compact version of the Llama series, suitable for tasks requiring lower computational resources.\\ \hline
\texttt{granite3-dense} & 1.6 & A model focused on creating dense textual representations for improved semantic understanding.\\ \hline
\texttt{smollm2} & 1.8 & A compact language model tailored for efficient performance in text analysis and generation tasks.\\ \hline
\end{tabular}
\end{table}

First, the ability to correctly identify errors in sentences was assessed, thereby measuring the true positive rate. Second was the number of false positives, representing instances of the model erroneously flagging correct text as incorrect. Third, the number of false negatives was quantified. 

\begin{figure*}[h!]
  \centering
  \includegraphics[width=\textwidth]{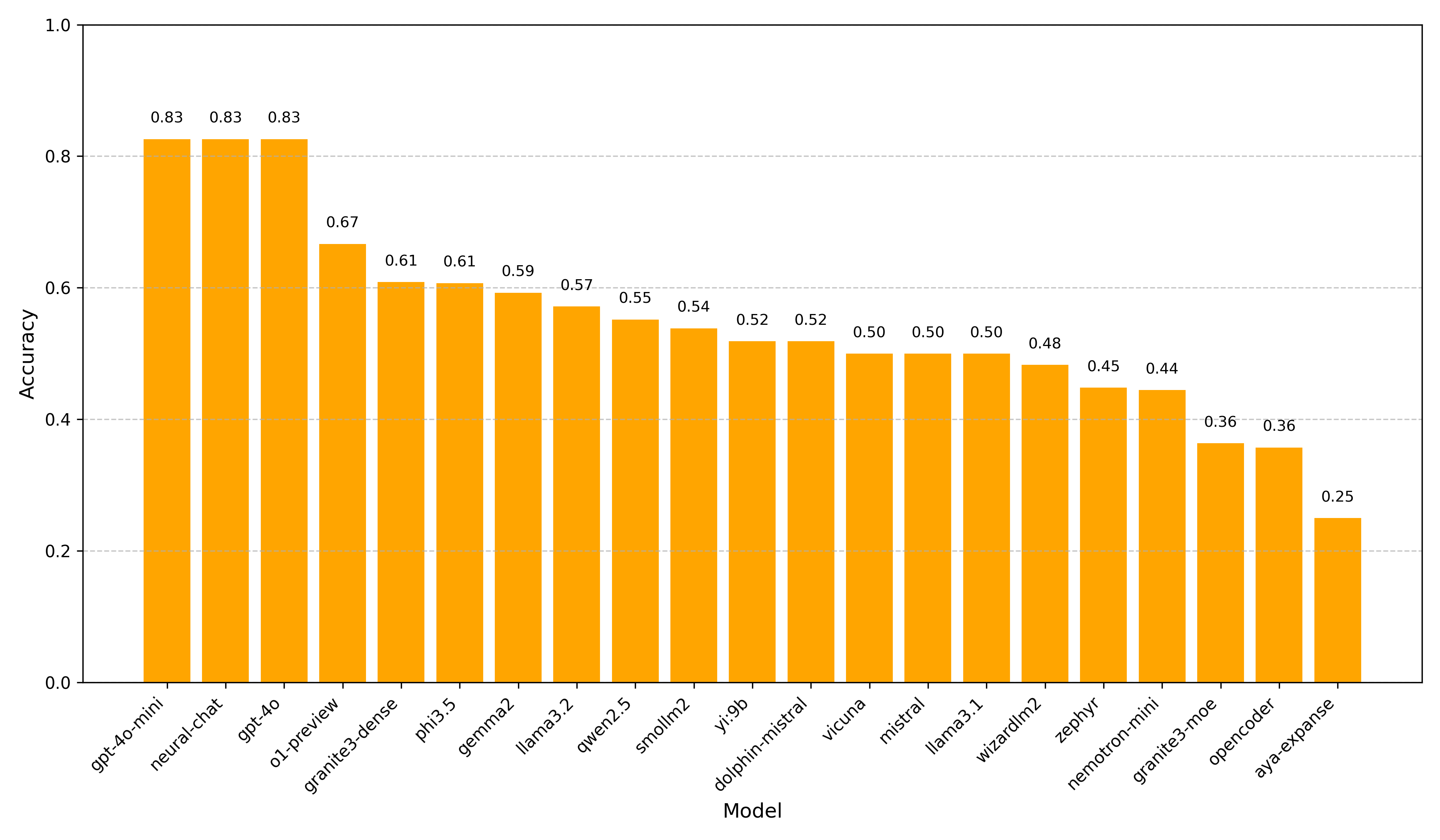}
  \caption[Accuracy per Model]{Accuracy of LLMs in detecting grammatical errors. Higher accuracy was observed in models such as \texttt{gpt-4o-mini} and \texttt{phi3.5}, while others like \texttt{aya-expanse} showed lower performance in this metric.}
  \label{fig:accuracy}
\end{figure*}

\begin{figure*}[h!]
  \centering
  \includegraphics[width=\textwidth]{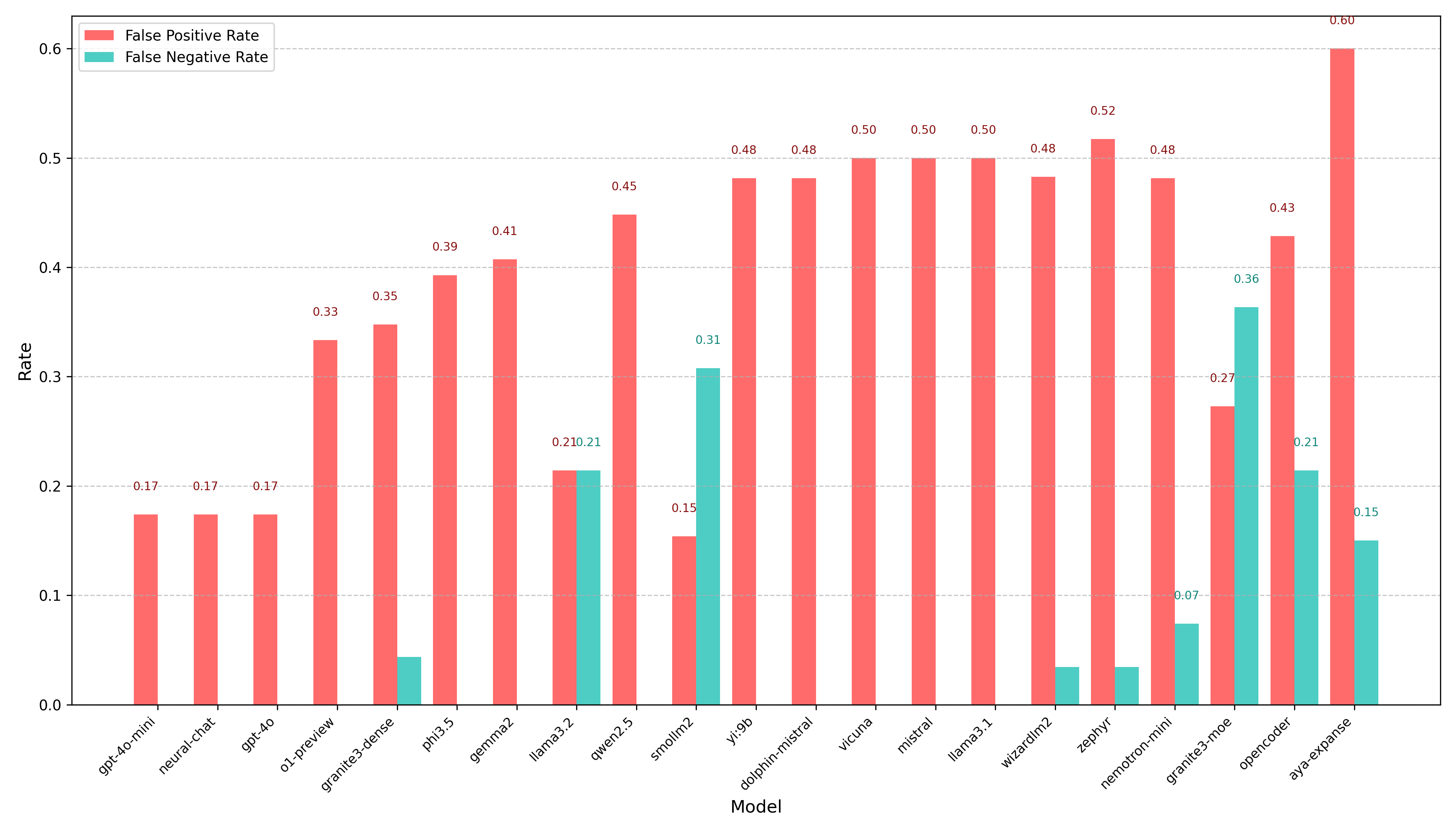}
  \caption[False Positive and Negative Rates]{False positive and false negative rates for each LLM, illustrating the models' accuracy in detecting grammatical errors.
  Models like \texttt{granite3-moe} achieved lower rates, indicating higher precision, while others exhibited higher error rates.}
  \label{fig:false_rates}
\end{figure*}

Eight sentences (see Appendix \ref{Appendix 1}) were selected from a corpus of modified English language learners' writing produced by Japanese university students, with sentences being modified to create a balance of correct, yet sometimes unnatural, and incorrect sentences and to ensure anonymity. The eight sentences were chosen because they reflect the complexity of language learners' writing-grammatically correct but unnatural sentences and sentences with multiple errors. Each model was given these eight sentences to analyze. The percentage of correct error detections for each model is presented in Figure~\ref{fig:accuracy}. The false positive and false negative rates were calculated and are visualized in Figure~\ref{fig:false_rates}, offering a comparative view across the different models.
 
This initial testing found gpt-4o and neural chat to be the most suitable models. Therefore, accuracy was further tested by analyzing a set of $200$ sentences (see Appendix \ref{Appendix 1})
using these two models. The $200$ sentences were randomly selected from the aforementioned corpus. Before performing the test, the $200$ sentences were independently rated by two authors, with discrepancies discussed and resolved, resulting in complete agreement on the presence of at least one error in $100$ sentences.

\begin{table}
\caption{Accuracy of Error Identification gpt-4o and Neural Chat.
    True Positives (TP), True Negatives (TN), False Positives (FP), False Negative (FN)}
    \centering
    \begin{tabular}{@{}lrrrr@{}}
    \toprule
        Model & TP & TN & FP & FN\\
        \midrule
       Gpt-4o  & 87 & 53 & 46 & 14\\
        Neural Chat & 82  & 59 & 41 & 18\\
        \bottomrule
    \end{tabular}
    \label{tab:accuracy}
\end{table}

\begin{table}
\caption{Precision, Recall, and F1 Score for gpt-4o and Neural Chat}
\centering
\begin{tabular}{@{}lccc@{}}
\toprule
Model & Precision & Recall & F1 Score \\
\midrule
Gpt-4o & 0.654 & 0.861 & 0.743 \\
Neural Chat & 0.667 & 0.820 & 0.735 \\
\bottomrule
\end{tabular}
\label{tab:metrics}
\end{table}

Table~\ref{tab:accuracy} presents the outcome counts used in classification metrics, including correct and incorrect predictions for both positive and negative classes. Using these values, precision, recall, and F1 score were computed to provide a more finer-grained view of classification performance as shown in Table~\ref{tab:metrics}. Results indicate that both models performed comparably in terms of F1 score differing in $0.8$ percentage points, though \texttt{gpt-4o} demonstrated slightly higher recall while \texttt{neural-chat} achieved marginally better precision.

\subsection{Appropriacy of Feedback Messages}

In addition to the percentage of true positives, a key aspect of performance is the quality of dynamic feedback. Therefore, the feedback provided for true positives by each model was evaluated using the three criteria of consistency, gradation and resolution, which are detailed here:

\begin{itemize}
    \item Consistency: the feedback focuses on the same error throughout all four feedback iterations in the event more than one error exists.
    \item Gradation: the feedback gradually becomes more explicit. 
    \item Resolution: where necessary the fourth iteration resolves the target error and improves the sentence.
\end{itemize}

When all three of these criteria were met, the feedback was evaluated as usable; when any of these three were absent, it was judged unusable. Gpt-4o provided usable feedback for 59 of its 87 true positives, while neural chat provided approximately 7 of 59.


False positives were further evaluated in terms of whether the feedback provided was deemed to be helpful for two reasons. First, grammatically correct sentences can be awkward and benefit from revision. Second, false positives have the potential to significantly negatively affect users. 
Therefore, the feedback on false positives provided by each model was evaluated for consistency, progression through graduated stages, and the inclusion of a resolution which improved on the original sentence. 
The results are as follows:

\begin{itemize}
    \item gpt-4o: 38 of 46 false positives generated usable feedback
    \item neural chat: 9 of 41 false positives generated usable feedback 
\end{itemize}

False negatives were not evaluated in further detail for several reasons. First, it is common pedagogical practice to ignore some errors as identifying all errors is overwhelming for learners. Second, teachers commonly miss some errors when providing feedback on the writing of learners. Finally, false negatives are unlikely to negatively impact user outcomes. 

\subsection{Suitability of Platform}

When evaluating the suitability of the platform, two key criteria are paramount: (1) scalability, which necessitates high processing speed, and (2) user-friendliness.

Performance testing was carried out to assess the responsiveness and scalability of the system under various usage conditions. 
A MacBook Pro 14 (Apple M2 Pro, 16GB) was utilized for running Ollama models offline, ensuring local processing capabilities for resource-intensive NLP tasks.

The dimensions examined were: average response time, response time stability, and scalability.

\begin{figure*}[h!]
  \centering
  \includegraphics[width=\textwidth]{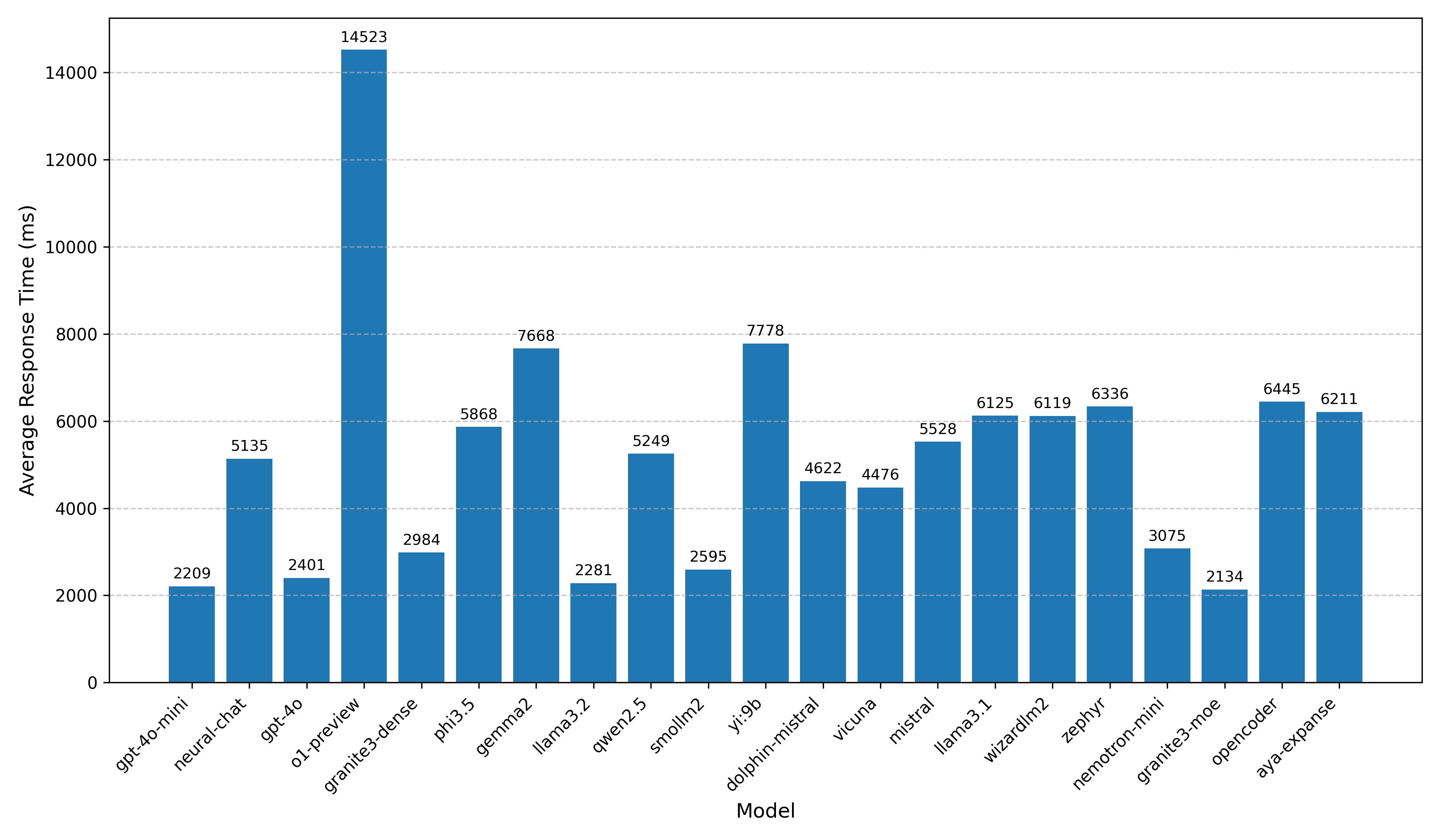}
  \caption[Average Response Time per Model]{A bar chart comparing the average response times of different LLMs, highlighting their suitability for real-time applications.
  Models such as \texttt{gpt-4o-mini} demonstrated the fastest response times, while others, such as \texttt{o1-preview}, showed significantly slower responses.}
  \label{fig:avg_response_time}
\end{figure*}

\begin{figure*}[h!]
  \centering
  \includegraphics[width=\textwidth]{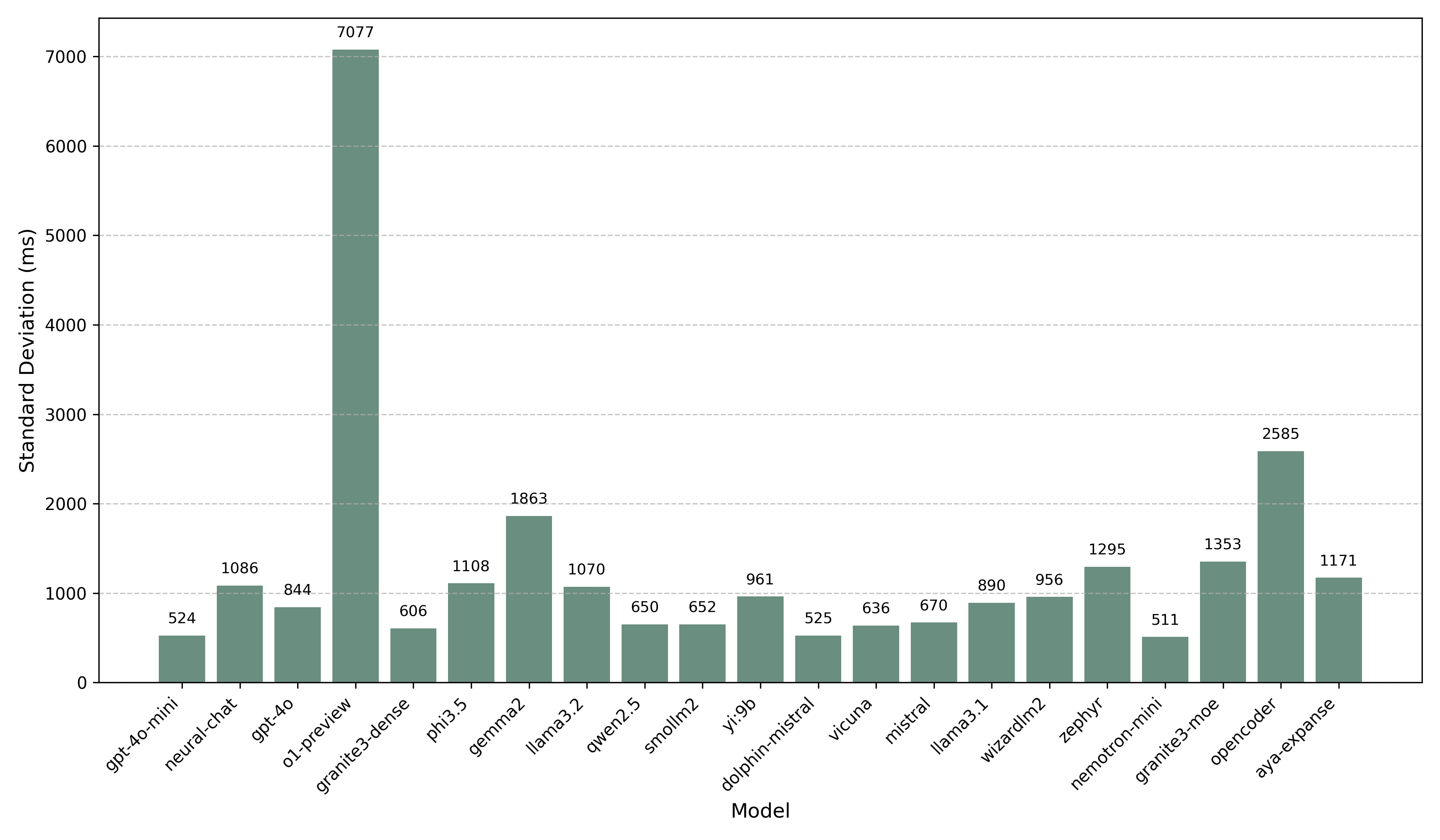}
  \caption[Response Time Stability]{The stability of response times across different LLMs, measured by the standard deviation.
  Models like \texttt{gpt-4o-mini} exhibited high stability, whereas \texttt{o1-preview} displayed significant variability in its response times.}
  \label{fig:response_stability}
\end{figure*}

Average response time was defined as the mean time taken by each model to process a request. Figure~\ref{fig:avg_response_time} presents these results, highlighting differences in processing efficiency across the models. Response time stability, indicated by the standard deviation of response times, is illustrated in Figure~\ref{fig:response_stability}. This metric provides insight into the consistency of model performance. Additionally, scalability was evaluated by observing the system’s ability to handle concurrent user requests without significant performance degradation.

\begin{figure*}[h!]
  \centering
  \includegraphics[width=\textwidth]{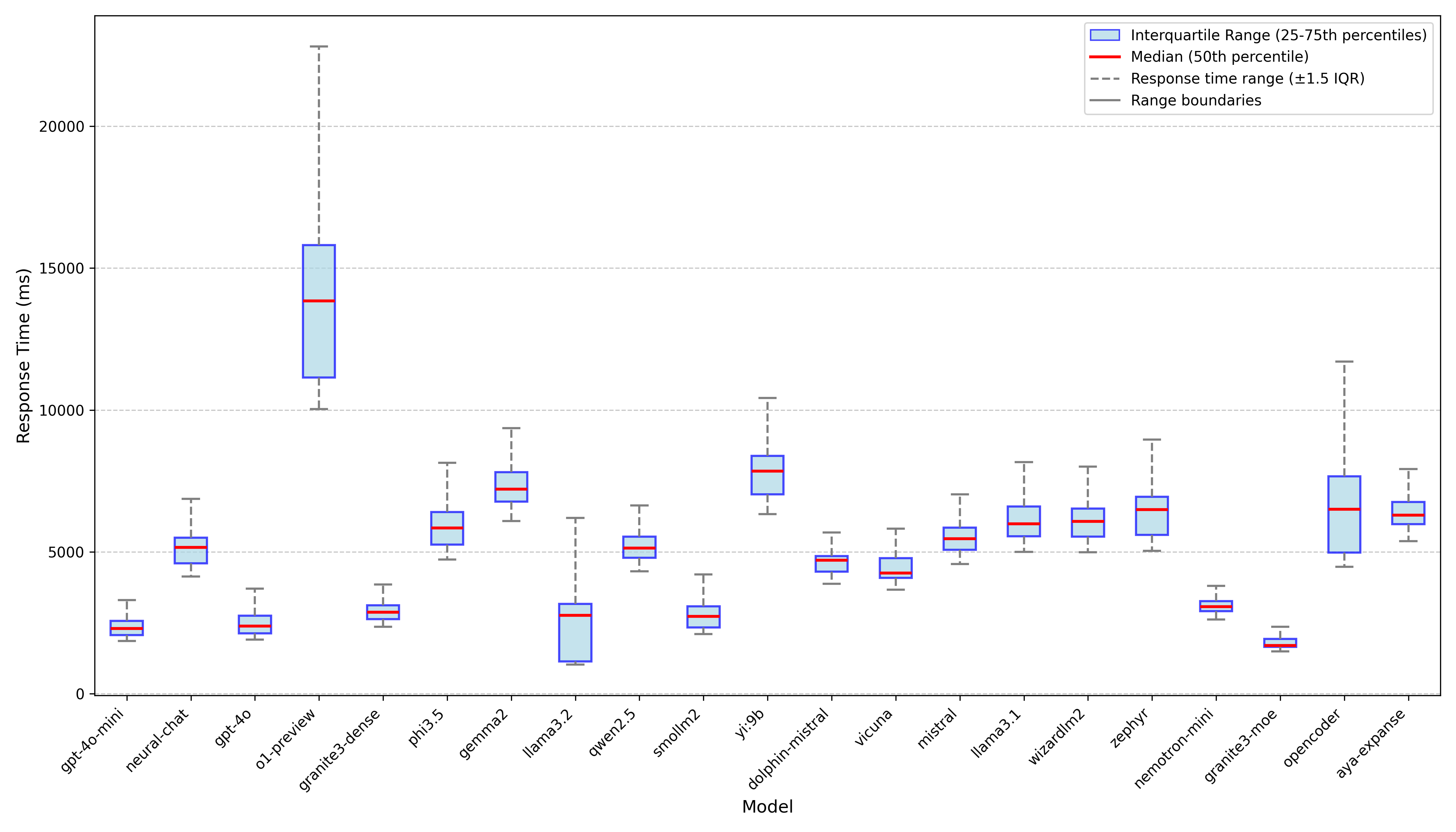}
  \caption[Response Time Percentiles]{Percentiles of response times for LLMs, showcasing the 25th, 50th (median), and 75th percentiles. This provides insights into the consistency of response times, with models like \texttt{gpt-4o-mini} demonstrating more predictable performance compared to models such as \texttt{o1-preview}.}
  \label{fig:percentiles}
\end{figure*}

These performance evaluations were supported by visual analytics and longitudinal tracking of user interaction data. Insights included metrics such as the average number of attempts required to correct a sentence and the effectiveness of the hierarchical hint system in guiding users toward accurate revisions. Figure~\ref{fig:percentiles} further outlines response time percentiles (25th, 50th, and 75th), offering a detailed view of model responsiveness under varying conditions.

Overall, the response times observed across the final two models were within acceptable thresholds to support real-time interaction and are sufficient to accommodate multiple concurrent users without compromising user experience or system responsiveness.

\section{Discussion}
\label{sec:discussion}

This section discusses results in terms of accuracy of error detection, appropriacy of feedback and scalability.

\subsection{Accuracy of Grammatical Error Identification}

The evaluation of $21$ models revealed substantial variation in performance, with models such as \texttt{gpt-4o-mini}, \texttt{phi3.5}, and \texttt{granite3-moe} demonstrating higher accuracy in identifying grammatical errors. 
From these, \texttt{gpt-4o} and \texttt{neural-chat} were selected for in-depth testing on a balanced dataset of $200$ sentences. 
Both models achieved similar F1 scores (approximately 0.74), with \texttt{gpt-4o} showing slightly higher recall and \texttt{neural-chat} slightly higher precision. 
These findings suggest that LLMs can be drawn upon to provide reliable grammatical error detection in extended learner writing. 
However, the presence of false positives remains a concern, particularly in pedagogical contexts where overcorrection may undermine learner confidence in the software. While precision levels are promising for formative feedback, further refinement is needed to improve the reliability of automated detection, especially for grammatically correct constructions.

\subsection{Pedagogical Appropriacy of Feedback}

The quality of feedback generated for true positive detections varied considerably between \texttt{gpt-4o} and \texttt{neural-chat}. \texttt{gpt-4o} produced usable feedback for approximately 68\% of true positives, meeting all three criteria of consistency, gradation, and resolution. 
In contrast, \texttt{neural-chat} produced usable feedback for fewer than 9\% of true positives. 
These results indicate that, while some LLMs are capable of delivering structured, pedagogically meaningful and graduated feedback, this capacity is not yet widespread across models. 
In terms of designing CDA, this is a key criterion for researchers when selecting which LLM to adopt.

False positives were also evaluated to determine whether the feedback provided could be considered helpful despite being technically grammatically correct. 
\texttt{gpt-4o} generated usable feedback in $38$ out of $46$ false positives, compared to $9$ out of $41$ for \texttt{neural-chat}. These findings underscore the importance of both the accuracy in detection and the pedagogical value of the feedback itself. 
Even in cases where the model misidentifies an error, the ability to offer coherent, graduated feedback that guides learners to a higher quality writing may go someway to mitigate negative user experiences.

False negatives were not subjected to qualitative analysis, as their impact on user experience was deemed minimal. Typically, during dynamic assessments and common pedagogic practices, some learner errors are intentionally left uncorrected to avoid overwhelming learners. It should also be acknowledged that human raters, including the expert raters in this study, similarly fail to identify all errors. For these reasons, false negatives were excluded from deeper analysis.

\subsection{Scalability and Usability of the Platform}

System performance was evaluated across three dimensions: average response time, response time stability, and scalability. Response times for the final selected models were within acceptable thresholds for real-time applications. Stability, as measured by standard deviation, was also sufficient to ensure predictable system behavior under load. These metrics indicate that the system is capable of supporting multiple concurrent users without significant degradation in responsiveness.

Longitudinal interaction data further supported these findings, showing consistent user engagement with the hint system and an acceptable number of revision attempts required to correct sentences. The response time percentiles confirmed that the majority of interactions occurred within a time frame conducive to effective learning.

Overall, we conclude that Dynawrite was able to meet the criteria for real-time deployment, offering a stable and responsive user experience; and thus LLMs can be used to implement CDA on learners' free writing. 

\subsection{Limitations}

The current implementation of DynaWrite, despite its robust architecture and broad functional capabilities, presents five key limitations that impact its scalability, accessibility, and overall comprehensiveness.

First, the system offers only partial coverage of available LLMs. While DynaWrite currently supports a range of models such as \texttt{GPT-4o}, \texttt{Llama3.2}, and \texttt{Vicuna}, it does not incorporate the full spectrum of models available in the field. Many specialized models—particularly those designed for low-resource languages, domain-specific applications, or fine-grained tasks—remain unexplored. Moreover, the fast-paced evolution of NLP technologies implies that numerous promising models will continue to emerge, which the current system is not yet equipped to accommodate.

Second, the system relies on infrastructure that may not be universally accessible. Some LLMs integrated into the platform, such as those accessed through the \texttt{OpenAI} \texttt{API}, depend on paid services, potentially limiting usability for institutions with budget constraints. Additionally, local deployment of models using the \texttt{Ollama} framework demands high-performance hardware, including \texttt{GPU}-enabled environments or advanced processors. This hardware dependency may present a significant barrier for smaller organizations or individual researchers.

Third, while the architecture of DynaWrite is modular and designed for scalability, effective large-scale deployment still requires substantial infrastructure enhancements. These include the integration of monitoring systems, load-balancing mechanisms, and optimized data storage solutions to ensure responsiveness and stability under high user loads.

Fourth, the internal analytics, though functional, are relatively basic. It currently tracks only fundamental metrics such as response times and error detection accuracy. More precise evaluation indicators are not yet incorporated into the assessment framework. These could include the linguistic complexity of hints, user revision behaviour, or the cognitive impact of feedback.

Finally, user experience design remains an underdeveloped area. Although the system has been validated for its technical functionality, limited attention has been given to user experience research. As a result, aspects such as interface usability and interaction flow, particularly for non-technical or novice users, require investigation to guide refinement to enhance accessibility and ease of use.

\subsection{Future Work}

Future development of DynaWrite will address its current constraints through a series of targeted enhancements. To overcome limitations in LLM coverage, future iterations will support a broader spectrum of models, including domain-specific models tailored to specialized tasks such as exam preparation. Mechanisms for automated benchmarking and dynamic evaluation will also be integrated to ensure that newly adopted models meet performance and pedagogical standards.

To improve accessibility and reduce dependency on high-performance hardware and paid APIs, development will focus on optimizing local deployment for resource-constrained devices and enabling offline functionality on mobile and edge platforms. These enhancements aim to make the system more inclusive and adaptable.

Scalability challenges will be addressed by migrating to \texttt{Kubernetes}-based container orchestration, coupled with real-time performance monitoring, auto-scaling, and advanced storage solutions for managing large-scale user interactions and datasets. These infrastructure improvements will support deployment in high-demand environments while ensuring system stability.

Limitations in evaluation metrics will be tackled through the inclusion of more granular assessment mechanisms, such as analyses of linguistic complexity, pedagogical effectiveness of feedback, and longitudinal tracking of learner progress. These metrics will offer deeper insights into learning outcomes and system efficacy.

Refinements to the feedback mechanism will include adaptive feedback strategies that align with individual learner profiles, multilingual support for error hints, and integration of cognitive learning theories to enhance instructional effectiveness.

Recognizing the current gap in user experience research, future work will also emphasize user-centric design improvements. These will include a simplified interface for non-technical users, customizable workflows for teachers and students, and the addition of visual aids and interactive onboarding tutorials to support new users.

Usability and user experience studies will be conducted to inform future interface and functionality refinements. Through structured feedback from both learners and educators, we aim to identify areas where the system can improve in terms of hint clarity, accessibility for non-technical users, and the intuitiveness of learner progress tracking. Insights gained from this process will inform iterative design enhancements and enhance the user experience. The tool is available at \url{dynawrite.org}.

Through these improvements, DynaWrite will remain a comprehensive, scalable, and pedagogically grounded platform that continues to meet the diverse needs of learners, educators, and researchers.

\appendices
\section{Test sentences}
\label{Appendix 1}
\begin{enumerate}
    \item All ages use email similarly.
    \item As age increases, the percentage of online game usage decreases.
    \item By this table I can find many things about how people use internet by age.
    \item First I am going to explain about email.
    \item In this category, there is no difference between age groups.
    \item Second is Online games.
    \item Teens play online games best than others.
    \item Third is downloading music and videos.
\end{enumerate}

The full set of $200$ test sentences is available upon request.



\begin{IEEEbiography}[{\includegraphics[width=1in,height=1.25in,clip,keepaspectratio]{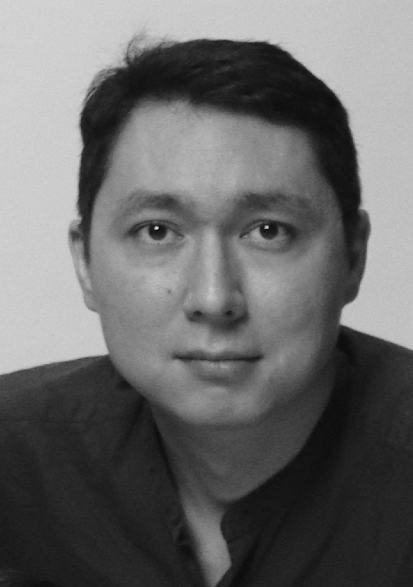}}]{Timur Jaganov} is pursuing a Master's in Information Technology and Project Management at the University of Aizu, Japan.
He holds a Bachelor's in Computer Science from the Techno-Economic Academy of Film and Television, Kazakhstan.
With extensive experience in software development, he has worked on designing scalable systems, implementing cloud-based solutions, and developing intelligent applications.
His research uses modern computational methods to enhance digital learning environments and optimize user interaction.

He is particularly interested in information systems, learning assistants, and distributed architectures.
Currently, he is conducting research in the School of Computer Science and Engineering at the University of Aizu, exploring innovative approaches to interactive education and knowledge management.
\end{IEEEbiography}

\begin{IEEEbiography}[{\includegraphics[width=1in,height=1.25in,clip,keepaspectratio]{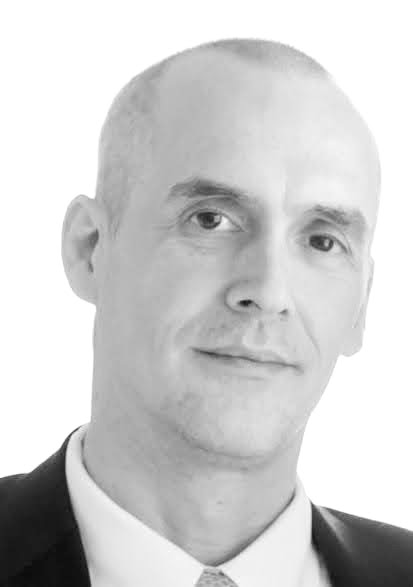}}]
{John Blake} received his Ph.D. in Applied Linguistics from the University of Aston, Birmingham, UK. He holds multiple postgraduate degrees, including an M.Sc. in Computer Science, an M.B.A., an M.Ed. in Applied Linguistics, an M.A. in Creative Writing, and an M.A. in Mathematics Education. 
He is a Chartered IT Professional and a professional member of the British Computer Society. His research primarily utilizes corpus linguistics to analyze texts and computational linguistics to develop pattern-searching tools and pipelines. Currently, he is a Professor in the School of Computer Science and Engineering at the University of Aizu, Japan. He also serves as an academic editor for PLOS One and reviews for multiple journals in the fields of natural language processing, linguistics and education. He has authored or co-authored over 80 papers.
\end{IEEEbiography}

\begin{IEEEbiography}[{\includegraphics[width=1in,height=1.25in,clip,keepaspectratio]{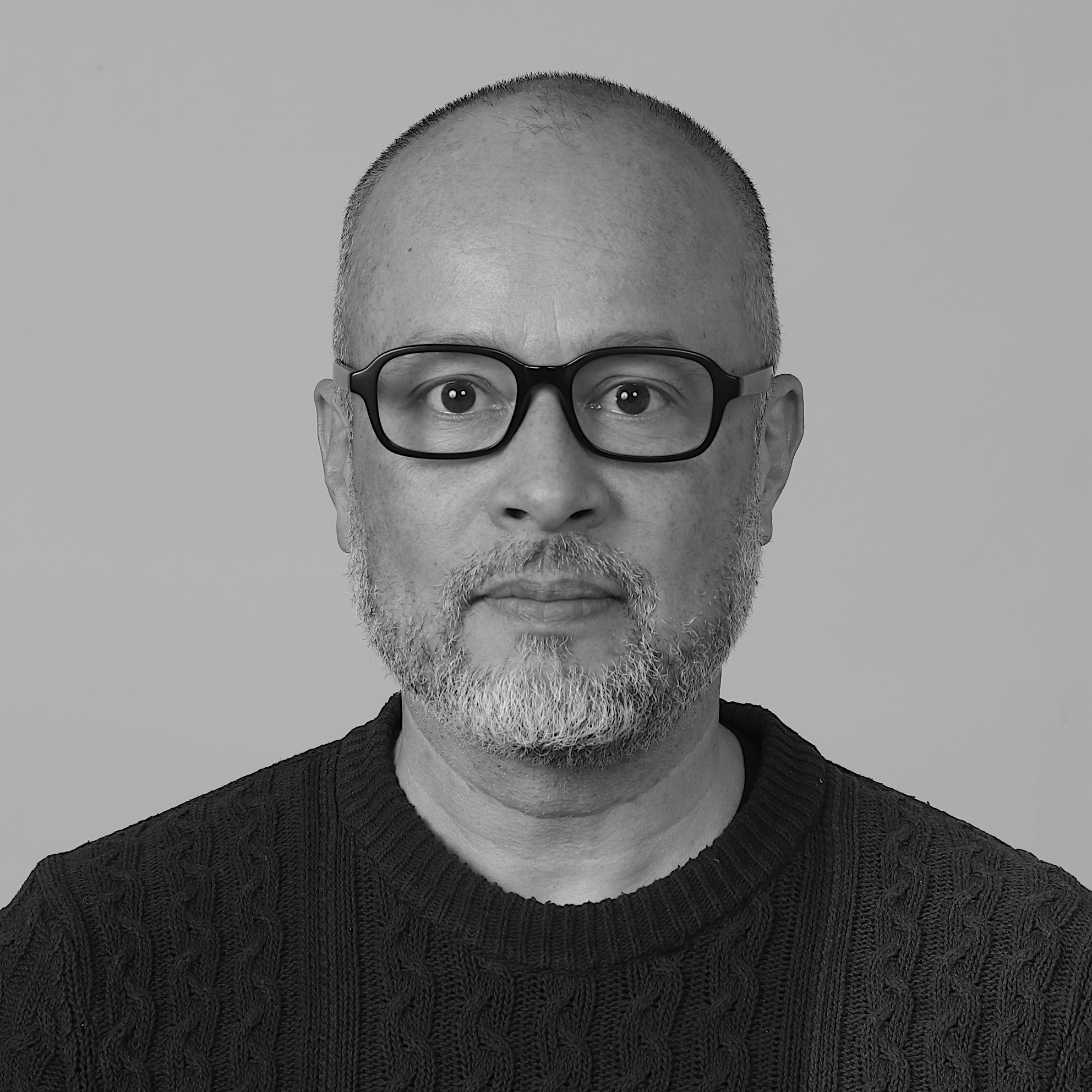}}]{JULI\'AN VILLEGAS} 
Senior Associate Professor at the Graduate School of Computer Science and Engineering, University of Aizu, Japan, with interests in speech intelligibility, music, and spatial sound.
\end{IEEEbiography}

\begin{IEEEbiography}[{\includegraphics[width=1in,height=1.25in,clip,keepaspectratio]{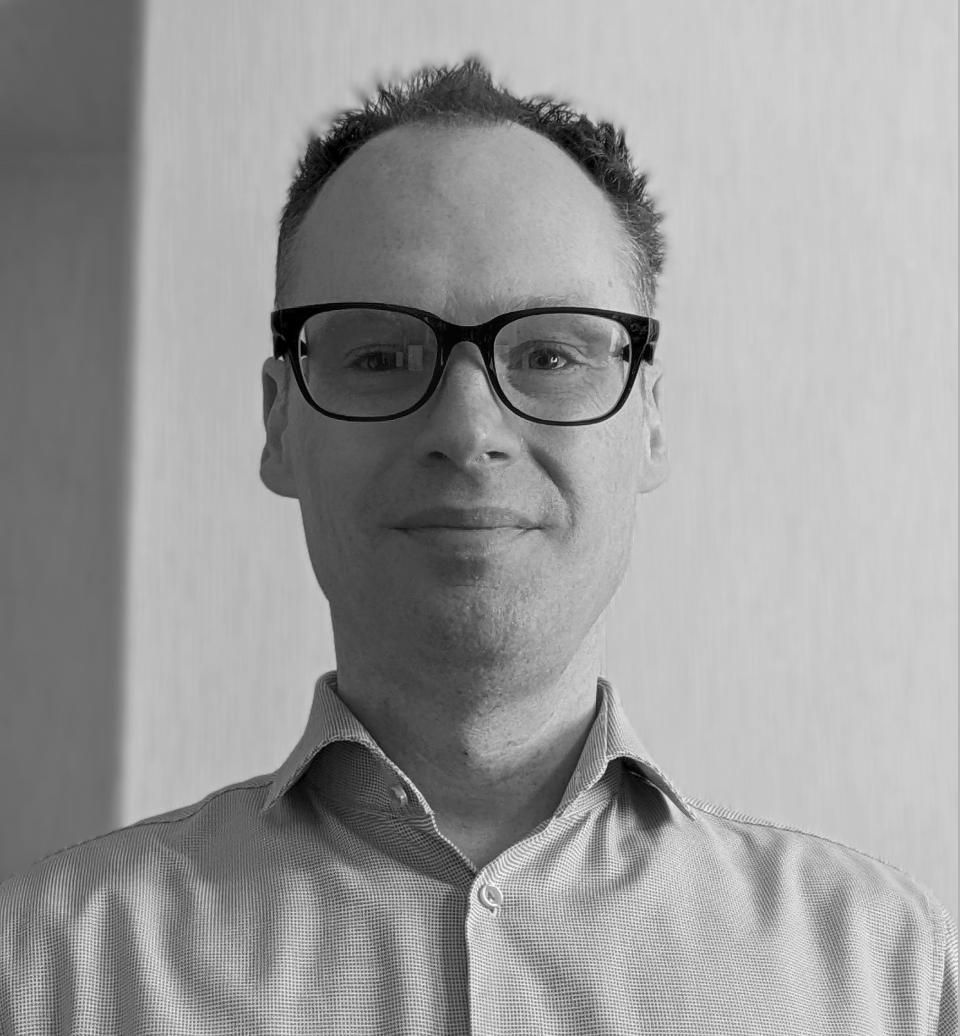}}]{Nicholas Carr} received his Ph.D in Education (TESOL) from Deakin University, Melbourne, Australia. He holds a Master of TESOL from Deakin University, a Diploma of Music Performance from Box Hill Institute, and Bachelor of Business from Swinburne University of Technology. 

His research interests include cognitive linguistics, concept-based-language instruction, sociocultural theory and computer assisted language learning. He currently serves as an Associate Professor in the School of Computer Science and Engineering at the University of Aizu, Japan. He also acts as a reviewer for multiple journals in the field of additional language learning and education. He has been published in several journals and edited volumes. 

\end{IEEEbiography}

\EOD

\end{document}